\documentclass[11pt,a4paper]{article}
\usepackage[hyperref]{acl2019}
\usepackage{times}
\usepackage{latexsym}

\usepackage{url}

\usepackage{hyperref}
\usepackage{amssymb}
\usepackage{wrapfig,lipsum,booktabs}
\usepackage{amsmath}
\usepackage{cleveref}
\usepackage{booktabs}
\usepackage{bm}
\usepackage{algorithm}
\usepackage{algpseudocode}
\usepackage{graphicx}
\usepackage{nccmath}
\usepackage{caption}
\usepackage{makecell}
\usepackage{subcaption}
\usepackage{tabularx}
\newcommand{\B}[1] {\boldsymbol{#1}}

\def\bu{{\B{u}}}

\DeclareMathOperator{\ner}{NER}
\DeclareMathOperator{\pos}{POS}
\DeclareMathOperator{\exactmatch}{EM}

\DeclareMathOperator{\BiLSTM}{\mbox{BiLSTM}}
\DeclareMathOperator{\softmax}{\mbox{softmax}}

\aclfinalcopy 

\title{SIM: A Slot-Independent Neural Model for Dialogue State Tracking}

\author{Chenguang Zhu \And Michael Zeng \\
\\
  Microsoft Speech and Dialogue Group, Redmond, WA, USA\\
  \texttt{\{chezhu, nzeng, xdh\}@microsoft.com} \\
  \And Xuedong Huang }

\date{}

\begin{document}
\maketitle

\begin{abstract}
Dialogue state tracking is an important component in task-oriented dialogue systems to identify users' goals and requests as a dialogue proceeds. However, as most previous models are dependent on dialogue slots, the model complexity soars when the number of slots increases. In this paper, we put forward a slot-independent neural model (SIM) to track dialogue states while keeping the model complexity invariant to the number of dialogue slots. The model utilizes attention mechanisms between user utterance and system actions. SIM achieves state-of-the-art results on WoZ and DSTC2 tasks, with only 20\% of the model size of previous models. 
\end{abstract}

\section{Introduction}
With the rapid development in deep learning, there is a recent boom of task-oriented dialogue systems in terms of both algorithms and datasets. The goal of task-oriented dialogue is to fulfill a user's requests such as booking hotels via communication in natural language. Due to the complexity and ambiguity of human language, previous systems have included semantic decoding \citep{nbt} to project natural language input into pre-defined dialogue states. These states are typically represented by slots and values: slots indicate the category of information and values specify the content of information. For instance, the user utterance ``can you help me find the address of any hotel in the south side of the city'' can be decoded as $inform(area, south)$ and $request(address)$, meaning that the user has specified the value \textit{south} for slot \textit{area} and requested another slot \textit{address}. 

Numerous methods have been put forward to decode a user's utterance into slot values. Some use hand-crafted features and domain-specific delexicalization methods to achieve strong performance \citep{henderson2014word, zilka2015incremental}. \citet{nbt} employs CNN and pretrained embeddings to further improve the state tracking accuracy. \citet{statnbt} extends this work by using two additional statistical update mechanisms. \citet{liu2018dialogue} uses human teaching and feedback to boost the state tracking performance. \citet{gl} utilizes both global and local attention mechanism in the proposed GLAD model which obtains state-of-the-art results on WoZ and DSTC2 datasets. However, most of these methods require slot-specific neural structures for accurate prediction. For example, \citet{gl} defines a parametrized local attention matrix for each slot. Slot-specific mechanisms become unwieldy when the dialogue task involves many topics and slots, as is typical in a complex conversational setting like product troubleshooting. Furthermore, due to the sparsity of labels, there may not be enough data to thoroughly train each slot-specific network structure. \citet{smdst,lsbt} both propose to remove the model's dependency on dialogue slots but there's no modification to the representation part, which could be crucial to textual understanding as we will show later.

To solve this problem, we need a state tracking model independent of dialogue slots. In other words, the network should depend on the semantic similarity between slots and utterance instead of slot-specific modules. To this end, we propose the Slot-Independent Model (SIM). Our model complexity does \textit{not} increase when the number of slots in dialogue tasks go up. Thus, SIM has many fewer parameters than existing dialogue state tracking models. To compensate for the exclusion of slot-specific parameters, we incorporate better feature representation of user utterance and dialogue states using syntactic information and convolutional neural networks (CNN). The refined representation, in addition to cross and self-attention mechanisms, make our model achieve even better performance than slot-specific models. For instance, on Wizard-of-Oz (WOZ) 2.0 dataset \citep{woz}, the SIM model obtains a joint-accuracy score of 89.5\%, 1.4\% higher than the previously best model GLAD, with only 22\% of the number of parameters. On DSTC2 dataset, SIM achieves comparable performance with previous best models with only 19\% of the model size.

\section{Problem Formulation}
As outlined in \citet{young2010hidden}, the dialogue state tracking task is formulated as follows: at each turn of dialogue, the user's utterance is semantically decoded into a set of slot-value pairs. There are two types of slots. \textit{Goal} slots indicate the category, e.g. area, food, and the values specify the constraint given by users for the category, e.g. South, Mediterranean. \textit{Request} slots refer to requests, and the value is the category that the user demands, e.g. phone, area. Each user's turn is thus decoded into \textit{turn goals} and \textit{turn requests}. Furthermore, to summarize the user's goals so far, the union of all previous turn goals up to the current turn is defined as \textit{joint goals}.

Similarly, the dialogue system's reply from the previous round is labeled with a set of slot-value pairs denoted as \textit{system actions}. The dialogue state tracking task requires models to predict turn goal and turn request given user's utterance and system actions from previous turns.

Formally, the \textit{ontology} of dialogue, $O$, consists of all possible slots $S$ and the set of values for each slot, $V(s), \forall s \in S$. Specifically, \textit{req} is the name for \textit{request} slot and its values include all the requestable category information.
The dialogue state tracking task is that, given the user's utterance in the $i$-th turn, $U$, and system actions from the $(i-1)$-th turn, $A=\{(s_1, v_1), ..., (s_q, v_q)\}$, where $s_j \in S, v_j \in V(s_j)$, the model should predict:
\begin{enumerate}
    \item Turn goals: $\{(s_1, v_1), ..., (s_b, v_b)\}$, where $s_j \in S, v_j \in V(s_j)$,
    \item Turn requests: $\{(req, v_1), ..., (req, v_t)\}$, where $v_j \in V(req)$.
\end{enumerate}

The joint goals at turn $i$ are then computed by taking the union of all the predicted turn goals from turn $1$ to turn $i$. 

Usually this prediction task is cast as a binary classification problem: for each slot-value pair $(s, v)$, determine whether it should be included in the predicted turn goals/requests. Namely, the model is to learn a mapping function $f(U, A, (s, v))\rightarrow\{0,1\}$.

\section{Slot-Independent Model}
To predict whether a slot-value pair should be included in the turn goals/requests, previous models \citep{nbt,gl} usually define network components for each slot $s\in S$. This can be cumbersome when the ontology is large, and it suffers from the insufficient data problem: the labelled data for a single slot may not suffice to effectively train the parameters for the slot-specific neural networks structure.

Therefore, we propose that in the classification process, the model needs to rely on the semantic similarity between the user's utterance and slot-value pair, with system action information. In other words, the model should have only a single global neural structure independent of slots. We heretofore refer to this model as Slot-Independent Model (SIM) for dialogue state tracking.

\subsection{Input Representation}
Suppose the user's utterance in the $i$-th turn contains $m$ words, $U=(w_1, w_2, ..., w_m)$. For each word $w_i$, we use GloVe word embedding $e_i$, character-CNN embedding $c_i$, Part-Of-Speech (POS) embedding $\pos_i$, Named-Entity-Recognition (NER) embedding $\ner_i$ and exact match feature $\exactmatch_i$. The POS and NER tags are extracted by spaCy and then mapped into a fixed-length vector. The exact matching feature has two bits, indicating whether a word and its lemma can be found in the slot-value pair representation, respectively. This is the first step to establish a semantic relationship between user utterance and slots. To summarize, we represent the user utterance as $X^U=\{\bu_1, \bu_2, ..., \bu_m\}\in \mathbb{R}^{m\times d_u}, \bu_i=[e_i; c_i; \pos_i; \ner_i; \exactmatch_i]$.

\begin{figure*}[htbp]
\includegraphics[scale=0.65,trim=0cm 0cm 0cm 3cm]{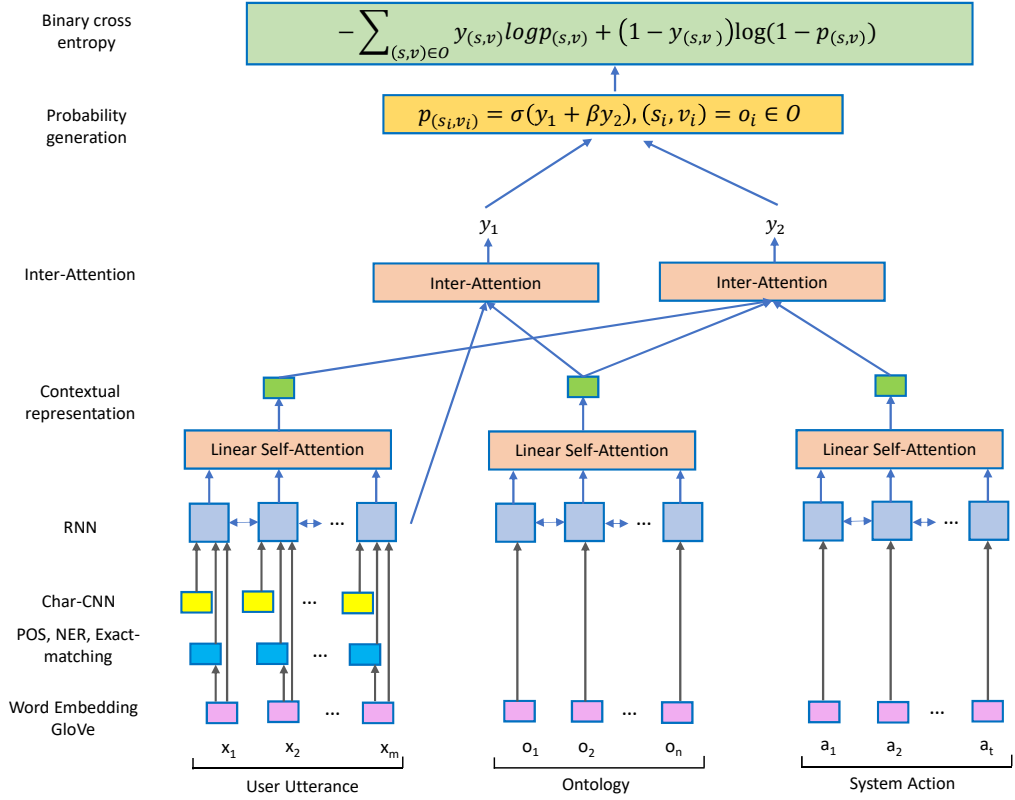}
\vspace{-4\baselineskip}
\caption{SIM model structure.}
\label{fig:model}
\end{figure*}

For each slot-value pair $(s, v)$ either in system action or in the ontology, we get its text representation by concatenating the contents of slot and value\footnote{To align with previous work, we prepend the word ``inform'' to goal slot.}. We use GloVe to embed each word in the text. Therefore, each slot-value pair in system actions is represented as $X^A\in \mathbb{R}^{a\times d}$ and each slot-value pair in ontology is represented as $X^O\in\mathbb{R}^{o\times d}$, where $a$ and $o$ is the number of words in the corresponding text.

\subsection{Contextual Representation}
To incorporate contextual information, we employ a bi-directional RNN layer on the input representation. For instance, for user utterance, 

\begin{equation}
    R^U = \BiLSTM{(X^U)} \in \mathbb{R}^{m \times d_{rnn}}
\end{equation}

We apply variational dropout \citep{vd} for RNN inputs, i.e. the dropout mask is shared over different timesteps. 

After RNN, we use linear self-attention to get a single summarization vector for user utterance, using weight vector $w\in \mathbb{R}^{d_{rnn}}$ and bias scalar $b$:

\begin{align}
    \alpha &= R^Uw + b \in \mathbb{R}^m \\
    p &= \softmax(\alpha) \in \mathbb{R}^m \\
    s^U &= (R^U)^Tp \in \mathbb{R}^{d_{rnn}} 
\end{align}

For each slot-value pair in the system actions and ontology, we conduct RNN and linear self-attention summarization in a similar way. As the slot-value pair input is not a sentence, we only keep the summarization vector $s^A \in \mathbb{R}^{d_{rnn}}$ and $s^O \in \mathbb{R}^{d_{rnn}}$ for each slot-value pair in system actions and ontology respectively.

\subsection{Inter-Attention}
To determine whether the current user utterance refers to a slot-value pair $(s, v)$ in the ontology, the model employs inter-attention between user utterance, system action and ontology. Similar to the framework in \citet{gl}, we employ two sources of interactions. 

The first is the semantic similarity between the user utterance, represented by embedding $R^U$ and each slot-value pair from ontology $(s, v)$, represented by embedding $s^O$. We linearly combine vectors in $R^U$ via the normalized inner product with $s^O$, which is then employed to compute the similarity score $y_1$:

\begin{align}
    \alpha &= R^Us^O \in \mathbb{R}^m \\
    p_1 &= \softmax(\alpha) \in \mathbb{R}^m \\
    q_1 &= (R^U)^Tp_1 \in \mathbb{R}^{d_{rnn}} \\
    y_1 & = w_1^T q_1 + b_1 \in \mathbb{R}
\end{align}

The second source involves the system actions. The reason is that if the system requested certain information in the previous round, it is very likely that the user will give answer in this round, and the answer may refer to the question, e.g. ``yes'' or ``no'' to the question. Thus, we first attend to system actions from user utterance and then combine with the ontology to get similarity score. Suppose there are $L$ slot-values pairs in the system actions from previous round\footnote{This includes a special sentinel action which refers to ignoring the system action.}, represented by $s_1^A, ..., s_L^A$:

\begin{align}
    p_2 &= \softmax(\{{s_j^A}^Ts^U\}_{j=1}^L) \in \mathbb{R}^L \\
    q_2 &= \sum_{j=1}^L{p_js_j^A} \in \mathbb{R}^{d_{rnn}} \\
    y_2 &= q_2^T s^O \in \mathbb{R}
\end{align}

The final similarity score between the user utterance and a slot-value pair $(s, v)$ from the ontology is a linear combination of $y_1$ and $y_2$ and normalized using sigmoid function.
\begin{equation}
    p_{(s, v)} = \sigma(y_1 + \beta y_2) \in \mathbb{R},
\end{equation}
where $\beta$ is a learned coefficient. The loss function is the sum of binary cross entropy over all slot-value pairs in the ontology:
\begin{align}
L(\theta)&=-\sum_{(s,v)\in O}y_{(s, v)}logp_{(s, v)}+\\
&(1-y_{(s, v)})log(1-p_{(s, v)}),
\end{align}
where $y_{(s, v)}\in \{0, 1\}$ is the ground truth.
We illustrate the model structure of SIM in \Cref{fig:model}.

\section{Experiment}
\label{exp}
\subsection{Dataset}
We evaluated our model on Wizard of Oz (WoZ) \citep{woz} and the second Dialogue System Technology Challenges \citep{dstc2}. Both tasks are for restaurant reservation and have slot-value pairs of both goal and request types. WoZ has 4 kinds of slots (\textit{area}, \textit{food}, \textit{price range}, \textit{request}) and 94 values in total. DSTC2 has an additional slot \textit{name} and 220 values in total. WoZ has 800 dialogues in the training and development set and 400 dialogues in the test set, while DSTC2 dataset consists of 2118 dialogues in the training and development set, and 1117 dialogues in the test set.

\subsection{Metrics}
We use accuracy on the joint goal and turn request as the evaluation metrics. Both are sets of slot-value pairs, so the predicted set must exactly match the answer to be judged as correct. For joint goals, if a later turn generates a slot-value pair where the slot has been specified in previous rounds, we replace the value with the latest content.

\begin{table*}[htbp]
\centering
\begin{tabular}{lcccc}
\toprule
 Model & \multicolumn{2}{c}{WoZ} & \multicolumn{2}{c}{DSTC2} \\
 & Joint goal & Turn request &  Joint goal & Turn request \\
 \midrule
SMDST & / & / & 70.3\% & /  \\
Delex. Model + Semantic Dictionary & 83.7\% & 87.6\% & 72.9\% & 95.7\%  \\
Neural Belief Tracker (NBT) & 84.2\% & 91.6\% & 73.4\% & 96.5\%  \\
LSBT & 85.5\% & / & / & /  \\
GLAD  & 88.1\% & 97.1\% & 74.5\% & \textbf{97.5\%} \\
SIM & \textbf{89.5\%} & \textbf{97.3\%} & \textbf{74.7\%} & 96.2\% \\
\bottomrule
\end{tabular}
\caption{Joint goal and turn request accuracies on WoZ and DSTC2 restaurant reservation datasets.} 
\label{table:mainresult}
\end{table*}

\subsection{Training Details}
We fix GloVe \citep{glove} as the word embedding matrix. The models are trained using ADAM optimizer \citep{adam} with an initial learning rate of 1e-3. The dimension of POS and NER embeddings are 12 and 8, respectively. In character-CNN, each character is embedded into a vector of length 50. The CNN window size is 3 and hidden size is 50. We apply a dropout rate of 0.1 for the input to each module. The hidden size of RNN is 125. 

During training, we pick the best model with highest joint goal score on development set and report the result on the test set.

For DSTC2, we adhere to the standard procedure to use the N-best list from the noisy ASR results for testing. The ASR results are very noisy. We experimented with several strategies and ended up using only the top result from the N-best list. The training and validation on DSTC2 are based on noise-free user utterance. The WoZ task does not have ASR results available, so we directly use noise-free user utterance.

\subsection{Baseline models and result}
We compare our model SIM with a number of baseline systems: delexicalization model \citep{woz,henderson2014word}, the neural belief tracker model (NBT) \citep{nbt}, global-locally self-attentive model GLAD \citep{gl}, large-scale belief tracking model LSBT \citep{lsbt} and scalable multi-domain dialogue state tracking model SMDST \citep{smdst}.

Table~\ref{table:mainresult} shows that, on WoZ dataset, SIM achieves a new state-of-the-art joint goal accuracy of 89.5\%, a significant improvement of 1.4\% over GLAD, and turn request accuracy of 97.3\%, 0.2\% above GLAD. On DSTC2 dataset, where noisy ASR results are used as user utterance during test, SIM obtains comparable results with GLAD. Furthermore, the better representation in SIM makes it significantly outperform previous slot-independent models LSBT and SMDST.

Furthermore, as SIM has no slot-specific neural network structures, its model size is much smaller than previous models. Table~\ref{table:modelsize} shows that, on WoZ and DSTC2 datasets, SIM model has the same number of parameters, which is only 23\% and 19\% of that in GLAD model.

\textbf{Ablation Study.} We conduct an ablation study of SIM on WoZ dataset. As shown in Table~\ref{table:ablation}, the additional utterance word features, including character, POS, NER and exact matching embeddings, can boost the performance by 2.4\% in joint goal accuracy. These features include POS, NER and exact match features. This indicates that for the dialogue state tracking task, syntactic information and text matching are very useful. Character-CNN captures sub-word level information and is effective in understanding spelling errors, hence it helps with 1.2\% in joint goal accuracy. Variational dropout is also beneficial, contributing 0.9\% to the joint goal accuracy, which shows the importance of uniform masking during dropout.

\begin{table}[h]
\centering
\begin{tabular}{l|cc}
\toprule
Model & WoZ & DSTC2 \\
\midrule
SIM & 1.47M & 1.47M \\
GLAD \citep{gl} & 6.41M & 7.69M\\
\bottomrule
\end{tabular}
\caption{Model size comparison between SIM and GLAD \citep{gl} on WoZ and DSTC2.}
\label{table:modelsize}
\end{table}

\begin{table}[h]
\centering
\begin{tabular}{lcc}
\toprule
Model & Joint Goal & Turn Request \\ 
\midrule
SIM & 89.5 & 97.3 \\
\quad --Var. dropout & 88.6 & 97.1 \\
\quad --Char. CNN & 88.3 & 97.0 \\
\quad --Utt. features & 87.1 & 97.1 \\
\bottomrule
\end{tabular}
\caption{Ablation study of SIM on WoZ. We pick the model with highest joint goal score on development set and report its performance on test set. } 
\label{table:ablation}
\end{table}

\section{Conclusion}
\label{conclusion}
In this paper, we propose a slot-independent neural model, SIM, to tackle the dialogue state tracking problem. Via incorporating better feature representations, SIM can effectively reduce the model complexity while still achieving superior or comparable results on various datasets, compared with previous models. 

For future work, we plan to design general slot-free dialogue state tracking models which can be adapted to different domains during inference time, given domain-specific ontology information. This will make the model more agile in real applications.

\section*{Acknowledgement}
We thank the anonymous reviewers for the insightful comments. We thank William Hinthorn for proof-reading our paper. 


\bibliography{dialogue}
\bibliographystyle{acl_natbib}

\end{document}